\documentclass[draftcls,12pt,onecolumn]{IEEEtran}

\usepackage{cite}
\usepackage{lineno}
\usepackage{xcolor}

\ifCLASSINFOpdf
  \usepackage[pdftex]{graphicx}
  \else
 
\fi

\usepackage{float}
\usepackage{lscape}
\usepackage{amsmath}
\usepackage{multirow}
\usepackage{color}
\usepackage{graphicx}
 \usepackage[colorlinks = true, linkcolor = blue, urlcolor = blue, citecolor = blue, anchorcolor = blue]{hyperref}

\begin{document}

\title{Towards a Collective Agenda on AI\\
for Earth Science Data Analysis}

\author{Devis Tuia,~\IEEEmembership{Senior Member,~IEEE}, Ribana Roscher,~\IEEEmembership{Member,~IEEE}, Jan Dirk Wegner, Nathan Jacobs,~\IEEEmembership{Senior Member,~IEEE}, Xiao Xiang Zhu,~\IEEEmembership{Senior Member,~IEEE}, Gustau Camps-Valls,~\IEEEmembership{Fellow,~IEEE}.
\thanks{DT was with Wageningen University, the Netherlands. He is now with Ecole Polytechnique F\'ed\'erale de Lausanne, Sion, Switzerland. E-mail: devis.tuia@epfl.ch (corresponding author). RR is with the  University of Bonn, Germany. JDW is with ETH Zurich  NJ is with the University of Kentucky (USA), XXZ is with the Technical University of Munich and the German Aerospace Center, GCV is with the Universitat de Val\`encia, Spain.}% 
\thanks{\noindent Digital Object Identifier  10.1109/MGRS.2020.3043504}
}

\markboth{IEEE GEOSCIENCE AND REMOTE SENSING MAGAZINE, Preprint, full version: 10.1109/MGRS.2020.3043504}{Tuia et al.: Towards a Collective Agenda on AI\\
for Earth Science Data Analysis}%
{}

\maketitle

\begin{abstract}
\textbf{This is the pre-acceptance version, to read the final version published in the Geoscience and Remote Sensing Magazine, please go to: \href{https://doi.org/10.1109/MGRS.2020.3043504}{10.1109/MGRS.2020.3043504}}\\
In the last years we have witnessed the fields of geosciences and remote sensing and artificial intelligence to become closer. Thanks to both the massive availability of observational data, improved simulations, and algorithmic advances, these disciplines have found common objectives and challenges to advance the modeling and understanding of the Earth system. 
Despite such great opportunities, we also observed a worrying tendency to remain in disciplinary comfort zones applying recent advances from artificial intelligence on well resolved remote sensing problems. Here we take a position on research directions where we think the interface between {these fields} will have the most impact and become potential game changers. In our declared agenda for AI on Earth sciences, we aim to inspire researchers, especially the younger generations, to tackle these challenges for a real advance of remote sensing and the geosciences. 
\end{abstract}


\section*{Introduction}
\label{sec:intro}

Artificial intelligence promises to change the way we do science. Nowadays it is widely accepted, almost a mantra, {that data along with faster computers and advanced machine learning algorithms can solve any data science problem.} Approaches issued from machine learning, computer vision, applied mathematics, or big data in general, are undoubtedly revolutionizing the way we tackle challenges in remote sensing and geosciences. This is particularly visible since deep learning has entered the arena~\cite{Rei19}: the promise of a technology able to process large amounts of data and to learn the complex structures of environmental processes, thus leading to an improved modeling, is making machine learning an unavoidable approach. For a multidisciplinary overview, see the recent book `Deep learning for the Earth Sciences'~\cite{CampsValls20dl4es}.

We are convinced that the rise of data-driven approaches in the geosciences is beneficial and will lead to important discoveries.  However, we want to raise awareness of pitfalls and fundamental questions that are largely obviated by the community.

A first risk is that of seeing everything as an opportunity of applying machine learning {and to then deploy massive technology regardless of whether such technology is necessary and adapted to the problem at hand}. 
Would that become a common practice, one would miss the 
opportunity of using (and improving) the new technology to tackle new challenges which were impossible before. For example, remote sensing is a data hungry discipline that has embraced machine learning very early: after an exploration phase (see the review papers~\cite{Zhu17,Aud19,Yuan20}), we see now the need for an impulse to embrace the technology to unlock new, difficult problems
, which in turn will \emph{create value for these geo-spatial data}. Part of the concepts presented below will sketch some of these directions (see Table~\ref{tab:overview}): first, the question on introducing reasoning in the modeling processing as a way to mimic cognitive processes about space and time (direction 1, page~\pageref{sec:reasoning}). Second, the need for exploring unconventional data modalities to capture the complexity of the visual world (direction 2, page~\pageref{sec:visualWorld}). And third, new ways of human interaction with remote sensing models, for instance via question answering as a way to retrieve image content on demand (direction 3, page~\pageref{sec:interactive}).

% !TEX root = main.tex

\begin{table*}[!t]
    \centering
    \caption{A summary of the six research directions presented in this position paper.}\label{tab:overview}
    \vspace{-0.25cm}
    {\scriptsize 
    \begin{tabular}{c|p{3.9cm}|p{0.5cm}|p{4.7cm}|p{5.7cm}|p{0.3cm}}
    \hline
         & In a nutshell& Refs. & Current issues & 10 years from now & Page \\\hline
         1 &Going beyond recognition towards induction, deduction, spatial and temporal reasoning, and structural inference.  & \cite{mou2019relation, mou2020era}  & Missing of or very limited benchmarks, novel tasks, as well as reasoning models, interpretability unsolved.  & Intelligent systems  linking meaningful transformation of entities, e.g. over space or time, and deriving knowledge, as the way people understand visual world and processes. & \pageref{sec:reasoning} \\ \hline
         2 & Think beyond the raster, consider all possible inputs and sources of supervision, in particular geotagged social media data.
         & \cite{aytar2017see,Perez-Rua_2019_CVPR,zamir2020robust}
         & {Presence of dataset biases, presence of label noise. Spatio-temporal mismatch between data sources. Scalability with increasing number of sources.} 
         &
         Systems that use a wide variety of sources to enable fine-grained understanding of the world, all with minimal human effort required for dataset building and system design.
         & \pageref{sec:visualWorld}\\ \hline
         3 &Query the world by asking questions to images, create descriptions. & \cite{Agr15,Lob19c} & Underexploited language part. Limited choice of thematic interactions. Lack of large scale infrastructure. & Visual search engines understanding questions about images, able to adapt to different types of requests and usable for everyone. & \pageref{sec:interactive}\\\hline
       
         4 & Make models learned with deep neural networks consistent with domain-specific knowledge like equations from physics. & \cite{karpatne2017physicsguided,ermon2017,Rei19,raissi2019,camps2018} & 
         {Networks' outputs are not physically consistent. 
         Networks are often used as emulators of simulations, but don't explore beyond current simulators constraints: they can't discover new physical rules.}
         & 
         {Systems trainable with much less data,  because they constrain output space via physical knowledge. Systems that learn new hypothesis for new science generation.}
         & \pageref{sec:physDL}\\\hline
           5 & Enhancing interpretability and explainability to understand processes in ML models in a better way. & \cite{roscher2020explainable,SamArXiv20} & 
         {Lack of human-understandable interpretation. Tendency towards confirmation bias (e.g. with attention maps)}
         & 
         {Models that are  more understandable, and therefore more reliable and trustworthy. Models that can be queried (and challenged) by humans about their inner reasoning}. & \pageref{sec:interpretable}\\\hline
         6 & Learn cause-effect relations, not just correlations, from observations and assumptions about the underlying generating process and system. &
         \cite{PearlMackenzie18,Peters18,PerezSuay19shsic,Runge19natcom}& 
         {Models can't work with unevenly sampled time series or non-stationary/noisy process. They extrapolate poorly.}
         &
         Machines that automatically blend domain knowledge, observational data and assumptions to learn the causal graph and generate causal narrative explanations of the problem. & \pageref{sec:causal}\\\hline
    \end{tabular}
    }
\end{table*}

A second pitfall is the blind faith in data science\footnote{Data science has been regrettably a misleading term. Shouldn't be replaced with `data for science' as some researchers suggest?  Science is about contrasting hypotheses, understanding physical phenomena and validating causal and explanatory models. If those objectives were achieved through data analysis, a new Science would be born.}. Driven by the impressive results obtained in machine learning and computer vision, it would be tempting to believe that everything can be solved with data and algorithms only. We believe that domain knowledge and model assumptions is of prime importance and that models must be challenged i) to respect the reality of the physical/biological/chemical processes governing the system under study and ii) to be accountable - by the transparency of their internal reasoning - of the decisions they lead to. This is important, especially when models are intended to be used for actual decision making and can affect balances of power or society changing decisions. 
{In the second part of the paper, we 
will present ideas along these directions  (see Table~\ref{tab:overview}) centered on the injection of domain knowledge, but for different purposes: first we will discuss physics-aware machine learning, which has the goal of using domain knowledge to restrict solution spaces of the models so that the outcome is physically plausible (direction 4, page~\pageref{sec:physDL}). This will grant physically consistency of the solutions and avoid aberrant outcomes that break physics (e.g. mass and energy conservation). Then, we will discuss how to obtain human understandable interpretations and explanations of the inner functioning of the models, in order to understand why and how models make decisions (direction 5, page~\pageref{sec:interpretable}). This has the advantage to make the model trustable and non-falsifiable and to avoid that the right conclusions are reached for the wrong reasons. Explainability also enhances the potential of testing new hypothesis and learn new scientific knowledge by the analysis of the model's functioning. These two first directions can be combined and use domain knowledge in different ways with different goals.
Transparency of the weights of the models is not absolutely necessary at this stage, as interpretations can be achieved by analyses of the inputs (e.g. LIME \cite{Rib16}) and physics awareness by modified loss functions. 
Yet, as argued before, science is about understanding the world we live in, not just to approximate it. We argue that without learning causal relations from observational data and assumptions, this ambitious goal of understanding the Earth system will not be possible  
(direction 6, page~\pageref{sec:causal}). 
In this case,  the learning of cause-effect relations is a mix of the previous ingredients, as domain knowledge is needed to design the model in such a way that it can reveal (maybe novel) cause-effect relations, which can be then explained using domain knowledge. Here transparency of the model is achieved by
construction, as the product is a self-explanatory causal graph. All three fields (explainability, hybrid modeling and causal inference) also
share in common the need of an  active and tight interaction between domain experts and computer scientists to make a decisive, non-incremental leap in Earth sciences.}
\vspace{0.3cm}

With this position paper we present six research directions {that we subjectively think hold particular promise for the future of Earth observation data analysis. In the following, we}  argue their potential and relevance, and provide some pointers to relevant resources. Our goal is to trigger curiosity and to foster successful research to truly advance the field of Earth sciences with Artificial Intelligence.

\section{Reasoning and human-machine dialogues}
 %!TEX root = main.tex

\label{sec:reasoning}

Current research at the interface between machine learning and remote sensing largely focuses on the direct recognition of materials, objects or on estimating geo-physical parameters. {Reasoning goes beyond the concept of recognition and aims at mimicking how people think and learn. It is centered around tasks such as induction, deduction, spatial and temporal reasoning, and structural inference \cite{lake_building_2016}. }

To date, only a few pioneering studies are published on reasoning for remote sensing tasks. 
In computer vision, reasoning is mostly interpreted as the capability to link meaningful transformations of entities over space or time. This is a fundamental property of intelligent species and also the way people understand visual data. Recently, papers implementing reasoning in CNNs started to appear: Santoro et al. proposed a relational reasoning network as a simple plug-and-play module to solve problems requiring the understanding of arbitrary relations between objects (ordering or comparisons of relative positions/sizes) and applied it to the problem of visual question answering (VQA) \cite{santoro2017simple} (more on VQA on direction 3, page~\pageref{sec:interactive}). A second pioneering work  concerns temporal relations in video sequences. When it comes to understand what takes place between two sampled video frames, humans can easily infer the temporal relations and transformations between  observations, unlike neural netowrks. In \cite{zhou2018temporal}, authors proposed a temporal relation network, which learns intuitive and interpretable common sense knowledge in videos.

\subsection*{Why Should Relational Reasoning Matter in Remote Sensing?}
Earth observation images carry strong spatial and temporal information, since each pixel is precisely referenced and connected to neighbors in space and time. When considering land processes (and in particular the geophysical ones), the relevant relationships can be learned by models. In \cite{mou2019relation}, authors  explicitly modeled long-range relations for semantic segmentation in aerial scenes. With the aim of increasing the representation capacity of a fully convolutional network (FCN), two tailored relation modules were used: {one} describing relationships between observations in convolved images and {another} producing relation-augmented feature representations. Given that convolutions operate by blending spatial and cross-channel information together, they captured relations in both spatial and channel domains.

\subsection*{Perspectives}

\begin{figure}[!t]
\centerline{\includegraphics[width=.95\columnwidth]{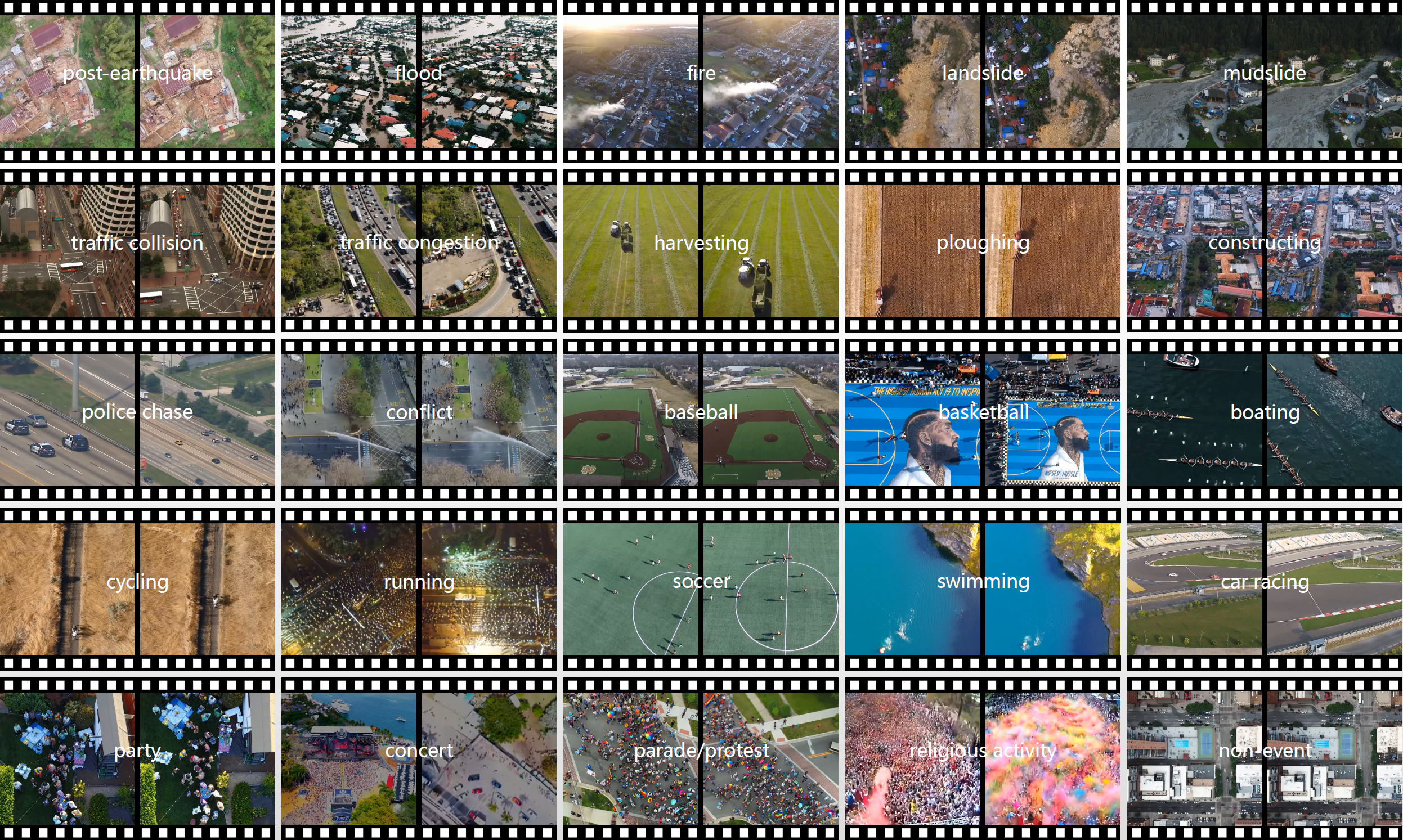}}
\vspace{-0.25cm}
\caption{Overview of the ERA dataset -- A benchmark for event recognition from aerial videos \cite{mou2020era}. For each class, the first (left) and last (right) frames of an example video are shown.}
\label{fig:reasoning}
\end{figure}

The work mentioned above showcases how spatial relational reasoning helps in improving semantic understanding of remote sensing images, and many other problems may also benefit from visual reasoning. One exciting example is temporal reasoning for the analysis of multitemporal data/aerial videos, e.g. for event recognition. 
{This is a new exciting field, where one is concerned by understanding complex events being imaged or filmed, such as cultural events, manifestations, or locating people in distress. Using reasoning enables to understand if a person on a roof during a flood is in actual need of help, or if a video of a crowd is related to a pacific or violent manifestation. This could be of interest to various stake holders including local authorities. To foster this research direction, authors in~\cite{mou2020era} introduce a dataset named ERA (Event Recognition in Aerial videos), consisting of three thousand  UAV videos manually annotated into  dozens of types of events  (Fig.~\ref{fig:reasoning}).}

Beyond videos, other interesting examples include VQA (see Direction 3, page ~\pageref{sec:interactive}), captioning \cite{Zhang_2019} and audiovisual reasoning, i.e., linking remote sensing images to in-situ audio signals~\cite{hu2020cross}. In the long run, we hope that reasoning Earth observation systems would be capable of deduce clues and make structural inference, in order to explain processes (see direction 5, page \pageref{sec:interpretable}) and understand causal structures in Earth Systems (see direction 6, page \pageref{sec:causal}).

\section{Extremely multi-modal remote sensing}
%!TEX root = main.tex

\label{sec:visualWorld}

Remote sensing is not restricted anymore to observation with airborne or satellite sensors. Nowadays, we can monitor our planet's health and status with social media data, socio-economic indicators, all kind of imagery, audio, and text, in addition to satellite imagery~\cite{Lef17}. {This direction raises several questions related to the importance of the different sources and their adequateness to specific tasks. In this research direction, we will discuss some of these aspects at the crossroads between sometimes extremely different data sources.} 
\subsection*{The traditional approach}

The first step in creating a traditional remote sensing system is to identify a property of interest, $y$ (e.g. land use or snow depth). We typically restrict to a set of locations, $l$, often described as a grid of points within a polygon, and times, $t$. We can formalize this as modeling a conditional distribution, $P(y|l,t)$, over the property, $y$, for a given set of locations and times.  In the traditional approach, one would resort to a pixel or object based classifier learning input-output relationships from a labeled  set of image pixels. The goal is  to provide an accurate estimate of the uncertainty over the property, $y$, and to have this generalize to new locations and times. 

The traditional approach is limited in several ways: (1) It requires a person to be able to label the training satellite imagery, either in the field or through manual image interpretation. This can be expensive and limiting, especially when asking the annotator to label difficult, fine-grained label spaces. (2) It is only able to make distinctions between phenomena that are easily visible from an overhead perspective. It cannot, for example, see inside buildings. (3) It is tightly coupled to the geographic region, the task, and source of data. This approach has led to a profusion of remote sensing papers use minor variations of the same computational methods.

\subsection*{Multimodal approaches with social media}

Social media data can be used to address our fundamental task, the estimation of $P(y|l,t)$. We begin by considering how {social media} can be incorporated as an input, by using $l$ and $t$ to query for nearby media content, and {then} how it can be used to expand the types of properties, $y$, that can be estimated. 

\begin{figure}[!t]
    \centering
    \includegraphics[width=\columnwidth]{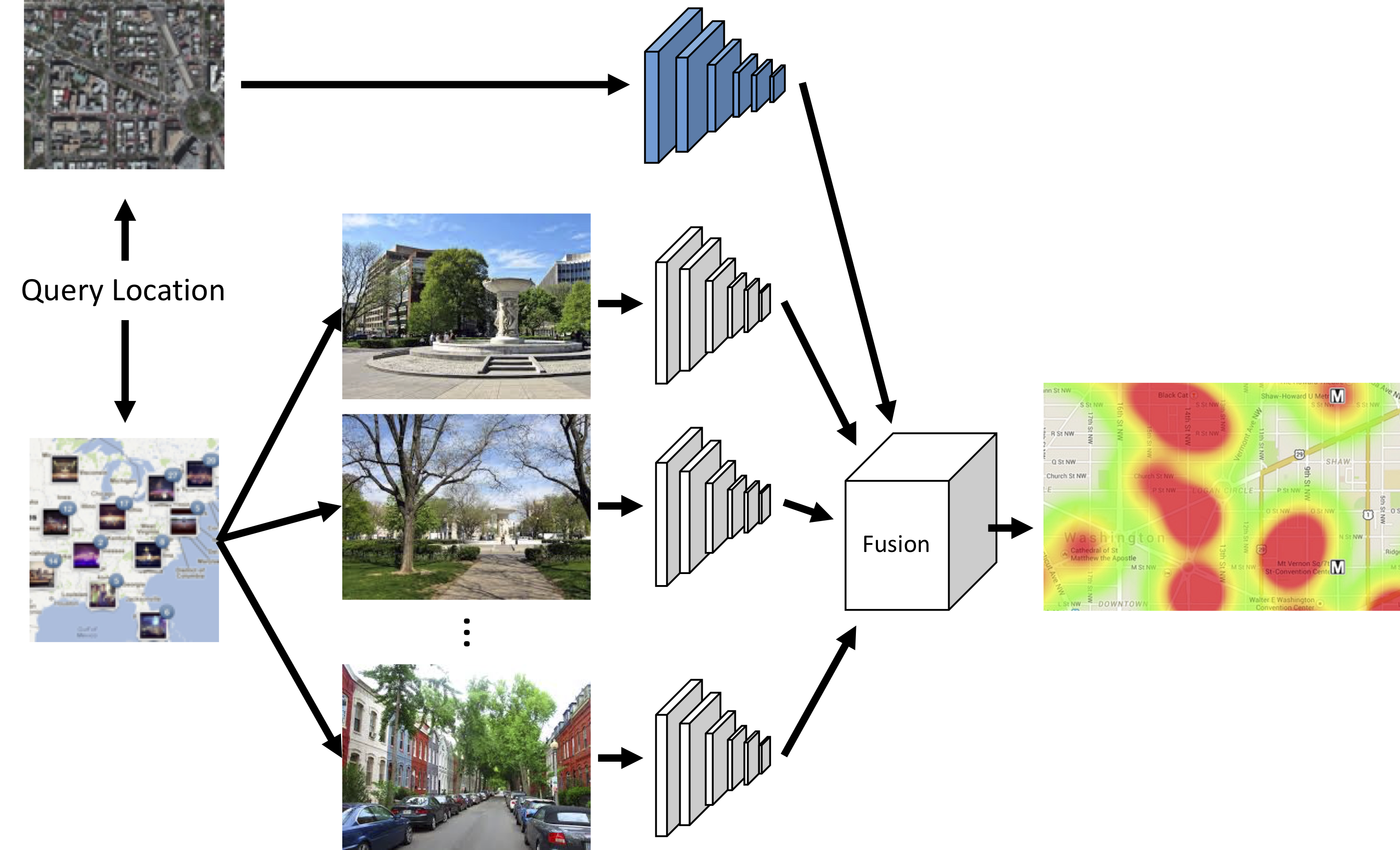}
    \caption{{Example of a system providing likelihood of presence of an object with multiple modalities. Ground and satellite images provide hard and weak (picture-based) labels respectively to create a heatmap of the presence of objects in urban areas. Location information is used to perform the fusion.}}
    \label{fig:jacobs}
\end{figure}

\paragraph{As an Alternative Input Modality} Traditional remote sensing largely ignores social media, and only uses $l$ and $t$ to index the image. However, today there are countless photographic, audio, and textual data being collected using cellphones. These data are often associated with the location and time of capture, making each a potentially useful source of information about the state of the world {(see Fig.~\ref{fig:jacobs} where a system based on both remote sensing and ground level images uses both modalities in synergy to provide likelihoods about the presence of objects)}. People make use of these data to make decisions on a daily basis. For example, when reading reviews and looking at photographs while trying to make travel plans. With the rapid advances in automatic interpretation of imagery, audio, and text we can start thinking of these data as potential inputs for remote sensing. Recent work~\cite{workman2017nearremote,kang_building_2018, srivastava2019understanding}
has begun to explore the use of social media imagery, especially for fine-grained landuse classification. A new methodological framework of information fusion with machine learning is actually emerging~\cite{salcedo2020machine}. They tend to use blackbox models to extract vector-valued features from the ground images and combine them with the remote sensing features. They also tend to ignore the rich geometric information the images contain. 

\paragraph{As a Source of Supervision} An untrained person can easily interpret a wide array of properties from a single social media object. The interpretation will be fine grained and include subjective properties such as dangerousness or scenicness, which are generally not visible in overhead images. With convolutional neural networks  approaching, and sometimes exceeding human-level performance for ground-level image interpretation, we can now consider using the output of CNNs as a semantic description of a given place and time. We can then use this description to train a remote sensing model, which might only take satellite imagery as input. {In this way, we can extend the information learned from social media to areas where these media are absent, and simultaneously reduce the need to manually annotate satellite images. }
This approach has been applied for a variety of tasks, including mapping scenes categories~\cite{workman2015localize}, and time-varying visual attributes~\cite{salem2020dynamic}. However, there remain significant issues to address, including:
\begin{itemize}

\item The data to be included: each source must add value, being correlated to the task and independent from each other. In an extremely multi-modal setting, {the volume and velocity of data acquired from each source cannot be directly controlled, since they depend on completely independent acquisition systems. Given that, it becomes important to understand how the spatial coverage and quality of each source varies. For example, clouds have a significant impact on optical imagery but do not affect SAR. Similarly, the coverage and quality of various social media sources depend on a variety of conditions, including: the proximity to tourist landmarks, population density, and differences in the culture of a given social network\footnote{We refer the interested reader to discussions about social media data quality in~\cite{Kit13,Var19c}.}. } Therefore, we must choose between satellite sources, but also social media types, social media platforms, and often must apply further filtering. For example, only including data collected from cellphone cameras~\cite{zhai2018geotemporal} or of certain types of scenes~\cite{workman2015localize}.

\item The quality of the matching between sources: it needs to accurately relate the ground and overhead perspectives. {The satellite/aerial to ground matching at the scene level is highly challenging due to the large semantic gap between the ground and overhead scene, but still resolvable. For example, the authors in \cite{lin_dual_2020} proposed an dual adversarial solution for unsupervised satellite/aerial to ground scene adaptation solution. However, it becomes very crucial when object level matching is concerned, e.g.,} when approaching automatic geolocalization \cite{shi2020optimal,shi2020looking}, and in particular when considering image synthesis \cite{Den18,regmi2019bridging,lu2020geometry}, where strong geometric models of the various modalities, with the ability to model uncertainty, are strongly needed.

\end{itemize}

\subsection*{Perspectives}
Considering social media data as extra sources for remote sensing analysis is gaining momentum. Besides classification and automatic geolocalization, using these additional data could unlock new applications, like modeling  soundscapes~\cite{salem2018soundscape}, landscape scenicness~\cite{workman2017scenic} {or place perception analysis~\cite{Zhu16,Zha18,Zha19}}. Moving even further, one could study phenomena closer to the consumer providing the data. For example, one could think of a system finding the most visually appealing driving route.

This will inevitably raise the question of dataset biases, since  social media data are personalized views of space: photographers tend to take pictures from places that are easy to reach, they are biased in the types of subjects they prefer and tend to take more pictures in the day time and in good weather conditions\footnote{See for instance the oversampling of particular photographic forms and scenes in the Instagram account insta\_repeat \href{https://www.instagram.com/insta_repeat/}{ (https://www.instagram.com/insta\_repeat/)}.}. This problem was recently considered when using observations collected by citizen scientists for species distribution mapping~\cite{BIRD2014144}{. In general, biases in learning models is a growing topic of study both in  the machine learning (see, for
example \cite{bahng2020learning,perez-suay_fair_2017}) and the social media (see reviews in \cite{Cihon16,MORSTATTER20171}) communities. There is wide room for such studies in remote sensing and biases issued by fusion in multimodal settings or hallucinations when using GANs.} 

Using social media also implies developing models that are robust to differences in appearance of classes, which becomes critical when predicting in new geographies or time moments. Since acquiring new labeled data is not always an option, one could envision to use these alternative sources as a form of weakly supervised training data, {or even as unsupervised supervisory signal for knowledge discovery, as authors in~\cite{Law19} proposed to explore the urban latent space of the streetscape of London}. Considering data acquired by autonomous vehicles and IoT devices will push this need even further, but will also unlock the potential of mapping on demand with extremely multi-modal remote sensing.

{Finally, further integration could make it possible to use social media as an early detection system for events, such as natural disasters~\cite{imr20}. For example, the social media imagery could be used to detect damaged structures or people in distress~\cite{chaudhary2020}. Such a system could even be used to cue satellite image acquisition over areas of interest based on image content, location data densities, or Tweets.}

\section{Interactive \& semantic machine learning}
\label{sec:interactive}
%!TEX root = main.tex

With the massive increase in  availability, remote sensing images are now used beyond scientific research. Firstly, images are available worldwide and with a high update rate. But they are also way more accepted by the general public: nobody is surprised anymore when shown a satellite view from Google maps;  consumer level drones can be used by virtually anyone and for all kinds of tasks: farmers monitoring crops, ecologists surveying animals or architects keeping track of construction sites are just a few examples. 

But despite the massive potential for image acquisition and updating, the usage of images remains static, in the sense that the images are mostly used for visualization, or at best to compute standard indices such as the NDVI, which is then assumed to represent vegetation status. Moreover, models answering specific needs of users are scarce and the more often limited to classical processing tasks (e.g. cars detection or land cover mapping) and cannot cover the variety of tasks different users could be interested in.
Another limitation is that end users rarely have the technical skills to design and run machine learning models, and would like to be able to receive an answer to a specific question of interest asked in natural language (e.g. in english).

Fortunately, a large variety of these questions boils down to the presence of objects, to counting or some kind of relational attribute (e.g. whether there was an increase of forest area or whether there are buildings in risk zones): a model able to pursue some kind of reasoning about the image content (see  direction 1, page~\pageref{sec:reasoning}), but taking into account a specific question (in english) by a user could open the door to a new type of interaction with remote sensing. Similarly to what search engines do on the Internet, a Remote Sensing Visual Question Answering (RSVQA~\cite{Lob19c}) engine could allow anyone (from scientists, to laymen and journalists) to retrieve the relevant information contained in the images.

Research in VQA is a vivid topic in computer vision~\cite{Agr15}, where it has had a lot of impact in creating systems to support vision impaired people in everyday tasks~\cite{Gur19}. A traditional VQA system in this context can indeed be used to help people when buying groceries, crossing the street, etc.

\subsection*{Dialogues between users and Earth observation images requires both remote sensing and natural language processing}

\begin{figure}
\centerline{\includegraphics[width=8cm]{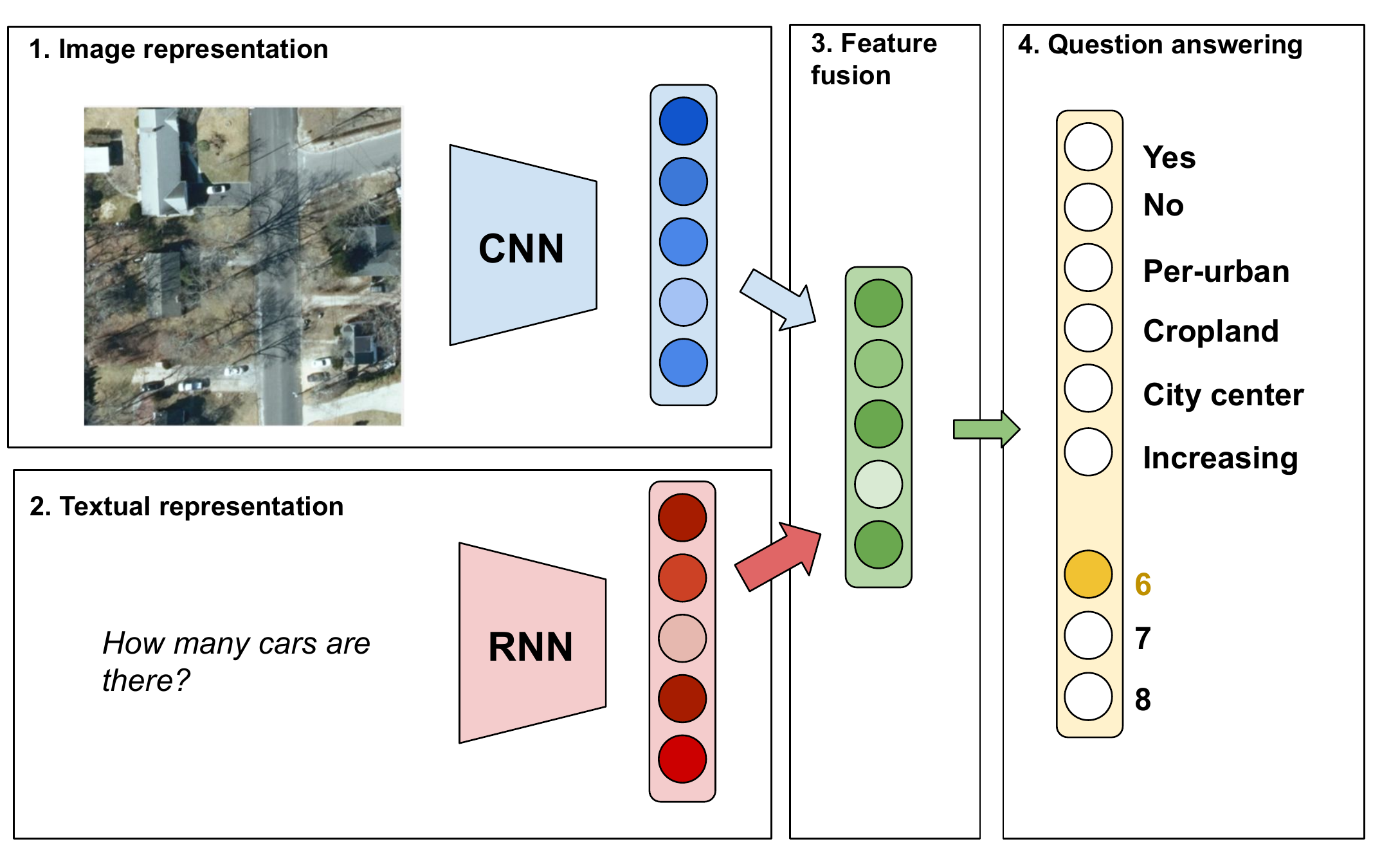}}
\vspace{-0.25cm}
    \caption{Example of a RSVQA system (modified from~\cite{Lob19c}): a remote sensing image (left, top) and a question in natural language (left, bottom) pair enter two source-specific neural nets, and outputting a vector representing their information; (middle) both vectors are combined and become the input of (right) a classifier that outputs possible answers as separate classes.
}\label{fig:rsvqa}
\end{figure}

\begin{figure}
\includegraphics[width=\columnwidth]{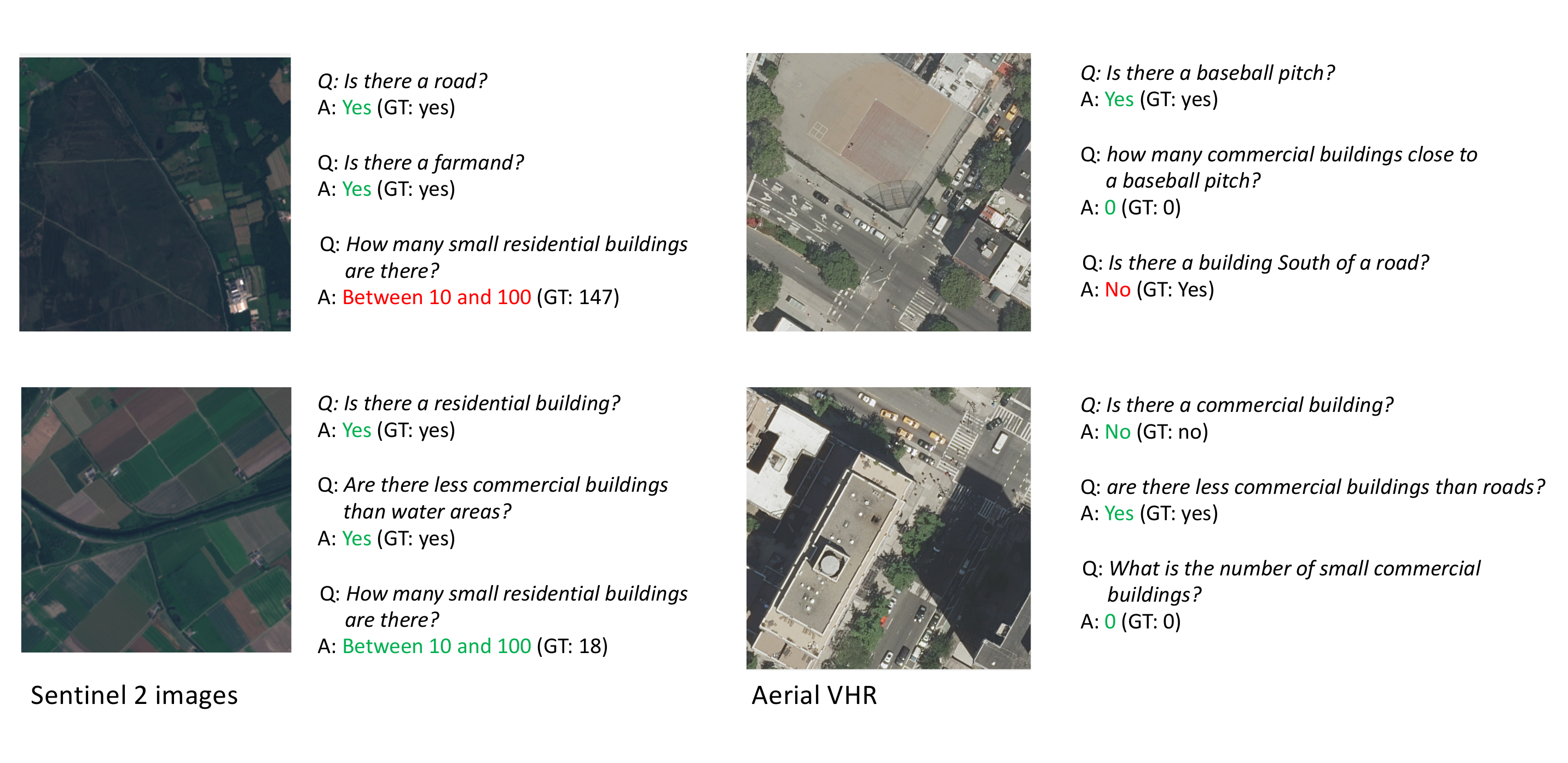}
\vspace{-0.5cm}
    \caption{Predictions of the RSVQA system on two Sentinel and aerial images, respectively, and different questions each. The same model is used to answer all questions related to one resolution imagery.}\label{fig:rsvqaEx}
\end{figure}

In remote sensing, the first VQA system was proposed in~\cite{Lob19c} and is summarized in Fig.~\ref{fig:rsvqa}. To become truly general-purpose, such model needs to be trained with a large quantity of data from several areas and different thematic objectives: in~\cite{Lob19c}  two models were designed, one for {Sentinel-2} data and another for subdecimeter resolution aerial images; the models were trained with large sets of image/answers pairs spanning tasks of classification, relative position reasoning and objects counting. Since a large quantity of labels was necessary, OpenStreetMap (OSM) vector data were used to automatically generate labels: following the CLEVR protocol~\cite{Jon17},  100 questions per image  involving objects occurring in the image (as informed by OSM) were generated. For each image/question pair, the answer (i.e. the label) was automatically obtained by querying OSM directly. Data and models are openly available at \url{https://rsvqa.sylvainlobry.com/}. Examples of predictions of the RSVQA model are reported in Fig.~\ref{fig:rsvqaEx} for both resolution images. Note that for a single image, several questions are possible and the same model is used to answer them.

\subsection*{Perspectives}

This first work opens a wide range of possibilities for a new line of research towards the next level of human/image interactions. Nonetheless, all the blocks of the model can (and must) be improved: for example, the automatic data generation has its flaws, especially due to the very simplistic language model used, for which new models from NLP could help improving the performance greatly. Also, less classical tasks (i.e. not reducible to classification, regression or detection) should be imagined, for instance allowing more complex output spaces: lessons learned from image captioning in remote sensing~\cite{Shi17} show that it is possible to move towards models that generating descriptions of the image content, which could be used, for instance, in image retrieval \cite{sumbul2020sdrsic}.

\section{Physics-aware machine learning}
%!TEX root = main.tex

\label{sec:physDL}

{Seen from the eyes of a practitioner, a major drawback of deep learning models
is that they} can lead to implausible results with scores that indicate high confidence in the outputs if no high-level constraints are imposed that check for consistency with theory. 

One possibility for compensating this shortcoming is integrating domain knowledge into the modeling procedure. Particularly in the environmental and geosciences, the laws of physics, chemistry, or biology govern the underlying processes and a lot of theory exists. An interesting direction of research is thus how to tightly couple machine learning and especially deep learning with physical laws. The hope is that this introduction of domain knowledge can help to reduce the manual labeling effort for supervised learning, counter dataset biases, reduce the influence of label noise, lead to good generalization capabilities, and eventually result in plausible outputs that adhere to the underlying physical principles. {An advantage of physics equations, and a difference to more generally explainable machine learning (see direction 5, page~\pageref{sec:interpretable}) and causal models (see direction 6, page~\pageref{sec:causal}), is that they can often be simplified into differentiable representations, which makes them amenable to backpropagation. In addition, emphasis in physics-consistent machine learning approaches for remote sensing is on modelling natural phenomena with higher accuracy, which is not necessarily the case for the two other research directions.} We will present some first ideas below, clustered into three lines of thought: constrained optimization, physics layers in deep neural networks, and encoding and learning differential equations. 
{A recent overview of the main families and approaches to the general field of the interaction between physics and machine learning for Earth observation is available in~\cite{CampsValls20physicsaware}.}

\subsection*{Constrained Optimization} 
\begin{figure}[t!]
\begin{center}
\includegraphics[width=7cm]{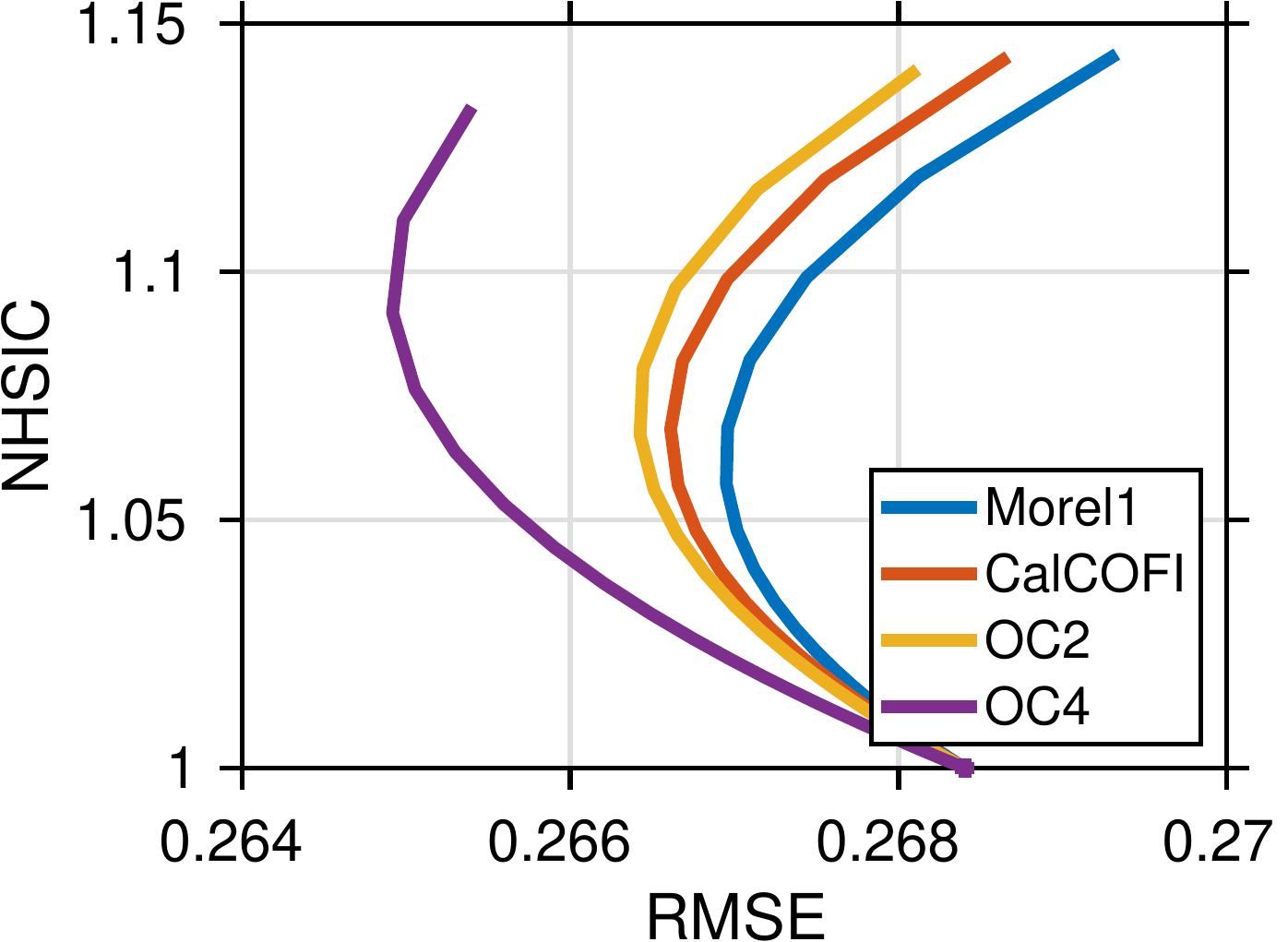}
\end{center} 
    \caption{
    {
    A standard family of hybrid modeling can be framed as a constrained optimization problem, where the physical rules are included as a particular form of regularizer \cite{von2019informed}. The fair kernel learning (FKL) \cite{perez-suay_fair_2017} method  enforces model predictions to be not only accurate but also {\em statistically dependent} on a physical model, simulations, or ancillary observations. In this example, we forced dependence of a data-driven model with respect to four standard ocean color parametric models (Morel1, CalCOFI 2-band linear, OC2 and OC4) and trained our constrained model to estimate ocean chlorophyll content from input radiances. We did so with increased dependency (as estimated by the NHSIC metric) between the machine learning and the physical  model. Results show that including the dependence regularizer (i.e. for higher NHSIC values) helps to reduce the RMSE and reveals that the OC2 and OC4 physical models in particular improve error and consistency of the data-driven model.
    }
    \label{fig:physicsawareexamples1}}
\end{figure}

A first idea to design physics-consistent machine learning approaches is imposing constraints in the loss function~\cite{karpatne2017,karpatne2017physicsguided}. Loss functions that encode the physical principles of a particular problem while using otherwise mostly unchanged model architectures can ensure that the learned model respects the laws of physics, {see Fig.~\ref{fig:physicsawareexamples1} for an example when including a dependence-based regularizer \cite{perez-suay_fair_2017}.} In addition, this strategy can significantly reduce the amount of necessary labels for training, up to practically zero in some cases~\cite{ermon2017}. 

Designing custom-tailored loss functions and possibly combining them with models that are trained on simulated data is another promising direction of research. However, this approach calls for very specific designs of loss functions that are not always straightforward and may simply not exist for many problems in remote sensing. For example, it seems very hard to design a corresponding loss function for semantic segmentation of cars in aerial images or detection of building facades in street-level panoramas, because the large intra-class variability of the appearances would require a very large set of constraints.

\subsection*{Physics Layers in Deep Neural Networks} 
\begin{figure}[t!]
\begin{center}
\includegraphics[width=8.5cm]{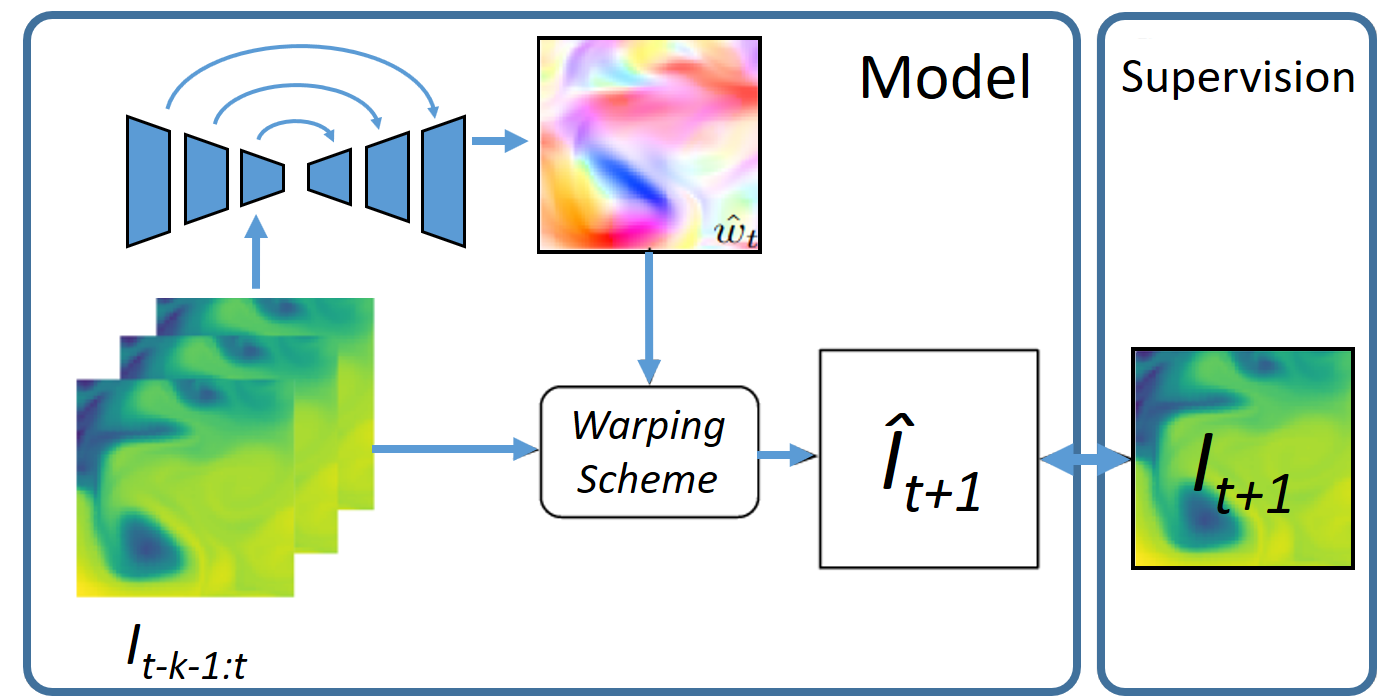}
\end{center} 
    \caption{
    {
    Another approach to hybrid ML modeling is that of including layers with physics motivation into a deep neural network, which are learned from data end-to-end. The shown architecture learns a motion field with a convolution-deconvolution net and the motion field is further processed with a warping physical model~\cite{bezenac2019}. The error is used to adjust the network weights, and after training the model can produce multiple time-step predictions recursively. {\sf \scriptsize 
    Credits: Figure adapted from \cite{bezenac2019}.}
    }
    \label{fig:physicsawareexamples2}}
\end{figure}

{An interesting idea to make use of well-established deep neural networks, but still learn and constrain the underlying physics, is adding additional layers that encode physics~\cite{bezenac2018,Rei19}, {see Fig.~\ref{fig:physicsawareexamples2} for an example}. General background knowledge gained from physics can be encoded in the deeper network layers. Together with a custom-tailored loss function, this approach enables end-to-end training of common deep networks that comply with physical constraints.}

{Although the idea of adding physical layers on top of common deep network architectures seems intuitive, implementing it for a wide range of remote sensing tasks is far from trivial. One idea could be to start with simplified versions that do not encode physics directly, but some related, simplified constraints, for example, for imposing maximum values for vegetation height mapping~\cite{lang2019}, tree stress~\cite{kalin2019}, or flood water depth~\cite{chaudhary2020}.} 

\subsection*{Encoding and Learning Differential Equations} Probably the biggest step towards deep neural networks that incorporate physics are so-called physics-informed neural networks (PINN) that directly encode nonlinear ordinary differential equations (ODE) and partial differential equations (PDE) in deep learning architectures while allowing for end-to-and training~\cite{raissi2018,raissi2019}. 
Instead of using standard network layers, the authors propose a framework to directly encode nonlinear differential equations in the network that is fully end-to-end trainable. 
This idea allows to learn yet unknown correlations and to come up with novel research hypothesis in a data-driven way, a central point also raised in the previous research direction on interpretability. Probabilistic models like Gaussian processes also allow encoding ODEs as a form of convolutional process~\cite{Svendsen20convproc}, and report additional advantages: besides the uncertainty quantification and propagation, they also learn the explicit form of the driving force and the ODE parameters, offering a solid ground for model understanding and interpretability {(see next research direction \ref{sec:interpretable}, on explainable machine learning)}.

\subsection*{Perspectives}
{Translated to remote sensing, physics-informed machine learning models allow to encode and learn radiative transfer equations and further physical laws like the backscattering of synthetic aperture radar (SAR) signal. Although directly starting with a full set of forward modeling equations like for simulation engines seems very hard, one could start with simplified versions and a subset of the most important components. 
Another idea would be encoding a simplified version of the change of spectral properties of vegetation as a function of seasonality. Similar to the warping model proposed in~\cite{bezenac2018}, one could encode change of spectral canopy properties in the infrared domain to ease domain transfer between summer and winter scenes.   
In the broader context of geosciences and climate sciences, learning ODEs/PDEs from observational data and simulations is the direct way to explain the problem and variable relations mechanistically, while still resorting to empirical data. The main challenges are the needed simplicity of the ODEs/PDEs that scientists can understand (so sparsity gets involved here) as well as validating the plausibility of such equations (so domain experts and computer scientists should work together).} 
{This links strongly physics-based deep learning to the explainable machine learning discussed in the next section.} 

\section{Interpretable and explainable machine learning}
%!TEX root = main.tex
\label{sec:interpretable}

\begin{figure}[t!]
\centering
\includegraphics[width=0.8\columnwidth]{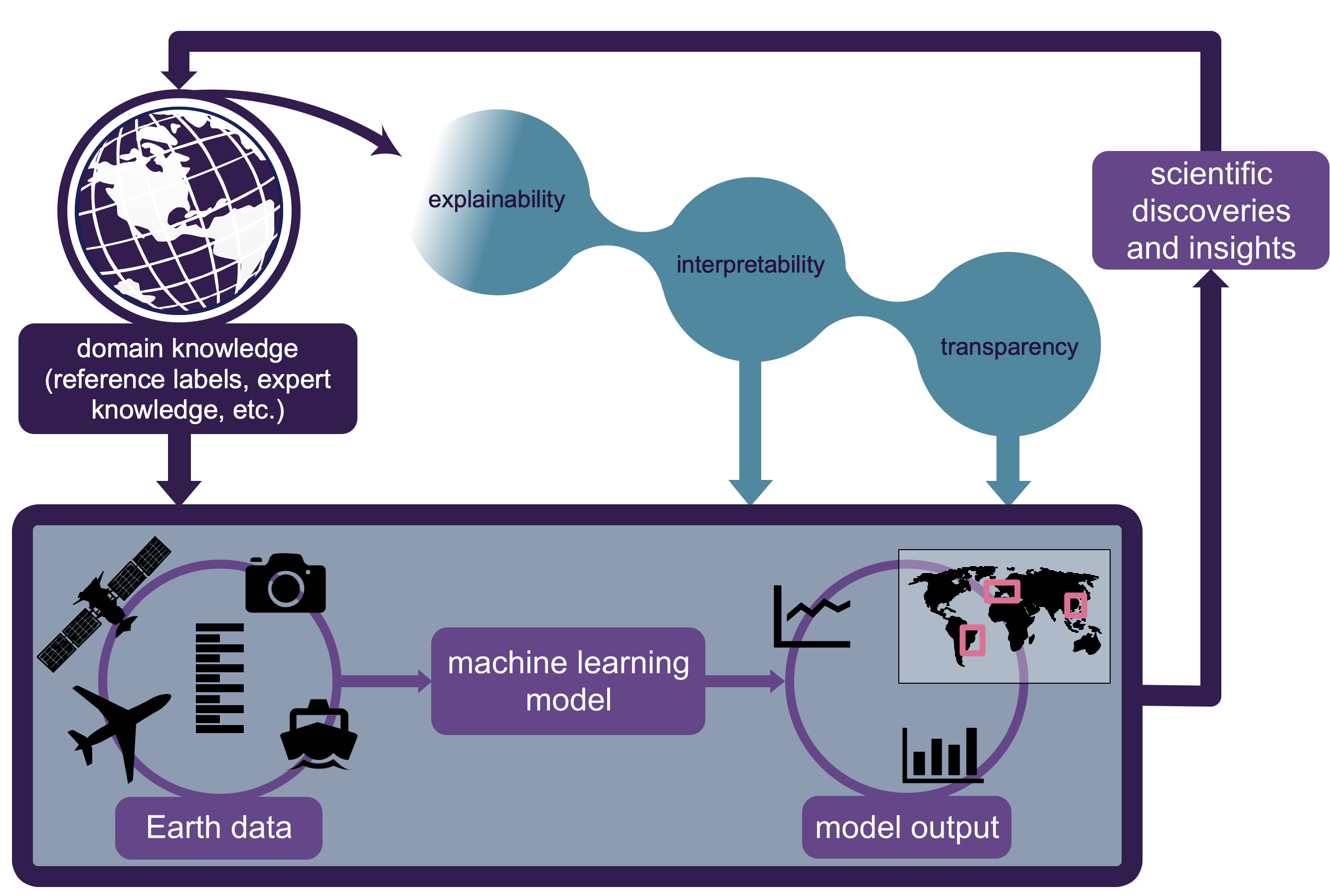}
\caption{{Explainable Machine learning can be used to gain scientific discoveries and insights by explaining a learned model and/or results (shown in the light gray box). Prerequisites are interpretability and potentially transparency that lead to scientific explanations when combined with domain knowledge. A feedback loop allows to extend and improve known domain knowledge. One application would be the derivation of improved definitions, for example for certain land use classes, which are currently only vaguely, incompletely, or not uniformly described.}}\label{fig:explain}
\end{figure}

Using machine learning for scientific applications aims at acquiring new scientific knowledge from observational data. {Additionally to the accuracy of the results, their scientific consistency, reliability, and explainability  are of central importance. A prerequisite to achieve those is to design models that can be challenged; in other words to create models whose inner functioning can be visualized, queried or interpreted.}
In this section, we will discuss the foundations of explainable AI (Fig.~\ref{fig:explain}), its exciting perspectives, and make links with physics-aware/informed machine learning that were discussed in the previous  section on physics-based ML (page~\pageref{sec:physDL}). 

\subsection*{From Transparency to Explainability}

Explainable machine learning has various definitions (see~\cite{roscher2020explainable}), but they all revolve around the properties of (1) transparency, (2) interpretability and (3) explainability: 
\begin{enumerate}
\item A transparent model allows us to access its components and ideally motivate why certain model components were chosen. This is in contrast to black box models as traditional neural networks, for which one could indeed write the mathematical relationships explicitly (they are transparent in this sense), but their complexity makes it inaccessible for users. 
\item An interpretable model counteracts the lack of transparency by presenting complex facts like the processes in a neural network in a space that can be understood by humans. 

Sorting by increasing interpretability power, such space can be made of localized image coordinates~\cite{Selvaraju_2017_ICCV}, semantic concepts \cite{Losch19} or understandable text~\cite{Park_2018_CVPR}. 
\item To achieve explainability, domain knowledge is exploited, which is used in combination with the  interpretable model and its components to understand, for example, why the model came to a certain decision.
Therefore, explanations become application-dependent and identical interpretations can lead to different explanations when linked to different domain knowledge.
\end{enumerate}

\subsection*{How Explainable AI may help remote sensing?}
Recently, many tools have been proposed for increasing interpretability and explainability when combined with domain knowledge~\cite{SamArXiv20}. Two major groups emerge: 
\begin{enumerate}
\item \emph{Post-hoc interpretability}. In this group, outcomes and decisions of the model are interpreted and explained by looking at the input. The most common visualizations for interpretations are heatmaps and prototypes. Heatmaps highlight parts of the input data that are prominent, important, or occlusion-sensitive. For example, they are created using the gradients flows in the neural network. Prototypes are optimized input data that, given a model, maximize the targeted output. Both these approaches help to understand what a model bases its decisions on, what influences the output, or what is a typical input for the learned input-output relationships. In all cases, attention must be paid to the confirmation bias. This is defined as the tendency to try to explain interpretations that are consistent with our existing knowledge, even if the explanation does not apply to the given case 

(see \cite{Ade18} for an example of overinterpretation of saliency maps).
\item \emph{Interpretability by design}. In this case, the model is inherently designed so that it can be interpreted.  Interpretability is achieved by representing model components or obtained latent variables in a way that they can be explained with knowledge from a certain application domain. 

{For example, units in hidden layers can be designed in such a way that the underlying factors of variation such as driving forces in Earth system data become disentangled and are captured in separate units. This could be seen, for example, by the fact that simple correlations exist between variations of the input and the activation of the neurons. 
Interesting applications of this idea are proposed in~\cite{Ye18}, where authors disentangle physical forces applying between objects in videos or in \cite{Marcos2019}, discussed below, for explaining human perception of beauty in landscapes.}
\end{enumerate}

\begin{figure*}[!t]
\includegraphics[width=\textwidth]{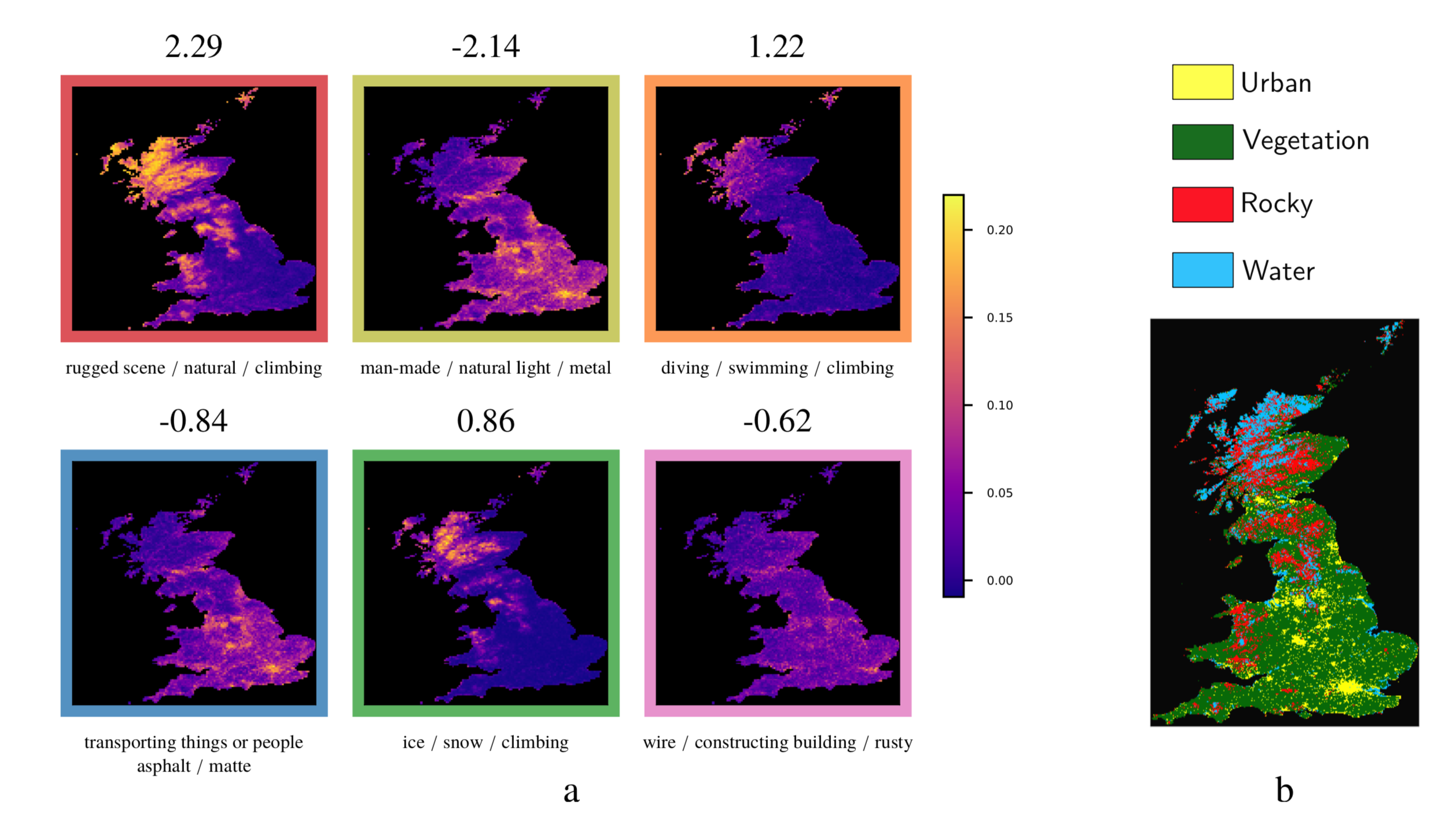}
\caption{{Explaining factors behind landscape beauty in the UK (from \cite{Mar20}): (a) Maps of landscape attributes contributing to the estimation of beauty in a series of landscape images; these maps are learned from ground based pictures and used to predict landscape attributes observed in the single images. Such attributes are then combined end-to-end in a neural network that predicts the beauty scores. The number above the single factor maps correspond to the contribution of the factor to landscape beauty on average. (b) Land cover map of the UK, used for visual comparison of the single factor maps learned automatically.}\label{fig:marcos}}
\end{figure*}

To ensure scientific value of the output, interpretation tools can be used to check its reliability. Besides the inherently existing output score of the neural network, for example, visualizations of the processes within the neural network can be used to check whether correct decisions have been made for the wrong reasons (so-called Clever-Hans-effect, \cite{Lapuschkin2019}). This can be seen as an additional test for the reliability of the output, due to the fact that a high score of the network does not always mean a correct result. 
{In summary, these tools can increase confidence by improving traceability as estimates are generated, and reveal biases in the data through human-understandable visualization.}

\subsection*{Perspectives}
Explainable machine learning has so far received comparatively little attention in remote sensing, partly because of the {still predominant opinion that explainability is tightly coupled with the complexity of a model and thus,} an increase in explainability leads directly to a decrease in accuracy {(e.g., \cite{rudin2019stop}). In the meantime, however, several applications have shown that this is not the case anymore.}

Most approaches consider so far are post-hoc interpretations, but first approaches considering interpretability by design are appearing.
In \cite{Marcos2019}, for example, the model is forced to predict human-interpretable concepts before predicting the final task {(Figure \ref{fig:marcos})}. Such approaches have the potential to provide both reliability checks and human understandable explanations, 
and could be used to go towards physics-explicit models as those discussed in the next section.

{As a further step, not only explanations can be allowed that are already understood with today's knowledge. New insights could be gained by using the explainable machine learning model and outcomes to formulate hypotheses for explanations that are not yet known to us. These hypotheses could then be tested with simulation software, for example, and confirmed or rejected. 
In order to show the potential, previous neural network approaches discover presumed scientific laws from observations without providing the complete underlying prior knowledge to the learning method  (e.g., \cite{Iten2018}). As illustrated in Fig.~\ref{fig:explain}, the next promising step would be to reveal new hypotheses from remote sensing data. This can, for example, be accomplished by searching patterns in interpretations, which can be assigned to novel discoveries and insights when combined with domain knowledge. This approach may point us to previously undiscovered spatio-temporal input-output relationships or data biases. For example, interpretation tools and the resulting insight could use how certain land-use classes are recognized to derive improved class definitions and help with a targeted acquisition of training data.}

\section{Learning cause-effect relations from data}
%!TEX root = main.tex
\label{sec:causal}

The Earth is a highly complex, dynamic, and networked system where very different physical, chemical and biological processes interact  in several spheres,  
and at diverse spatio-temporal scales. 
Despite the great predictive capabilities of current machine and deep learning methods, there is still little actual {\em learning}. Understanding is harder than predicting. Machine learning algorithms 
excel in fitting arbitrary functional data relations but do not have a clear notion of the underlying causal relations. Machine learning is far from problem understanding and even more from machine intelligence\footnote{See critical perspectives in the blogs by \href{https://medium.com/@GaryMarcus/the-deepest-problem-with-deep-learning-91c5991f5695}{Gary Marcus} and \href{https://medium.com/@mijordan3/artificial-intelligence-the-revolution-hasnt-happened-yet-5e1d5812e1e7}{Michael Jordan}, and the perspective papers~\cite{Rei19,Runge19natcom}}. 

Earth system data analysis aims to extract information from multivariate non-grided datasets, where missing data, nonlinearities and nonstationarities are present in the wild. Variables and physical processes are coupled in space and time, and (tele)connections can be of large-range, discontinuous are variant in strength and intensity. Addressing this problem will allow us to identify the right set of predictors, develop robust models and to avoid getting the right answer for the wrong reasons. The links with physics (direction 4, page \pageref{sec:physDL}) and  interpretability (direction 5, page \pageref{sec:interpretable}) are very strong.

\subsection*{Causality as the way forward}

{\em Causal inference} aims at discovering and explaining the causal structure of the  system~\cite{Peters18,ZhaSchSpiGly17,Runge19natcomm}. Very often, interventions in the system are not possible because of ethical, practical or economical reasons. Then {\em observational} causal inference comes into play to extract cause-effect relationships from multivariate datasets, going beyond the commonly adopted correlation approach, which merely captures associations between variables.

% Causality from observations
Today the science of ``causal inference'' \cite{Pearl2000,PearlMackenzie18} is fast advancing and, under reasonable assumptions, can unravel {\em causal} relations between two or more coupled variables even in the presence of non-linearities and non-stationarities, and even when time is not even involved. Several rigorous algorithms have been developed in the last decade that allow us to make inferences across multiple variables to discover plausible causal relations from observations. Causal inference is of course very relevant for the scientific endeavour, but it also has impactful practical implications. For example, learning causal structures allows us to build more parsimonious and robust models and that means faster, more fault-tolerant, and interpretable models.  

% A taxonomy of causal inference methods, made simple and standard
\subsection*{A taxonomy of causal discovery methods}

\begin{figure*}[t!]
\begin{center}
    \begin{tabular}{ccc}
(a) Nonlinear Granger causality & (b) Convergent-cross mapping & (c) Additive noise model \\
\includegraphics[height=4cm]{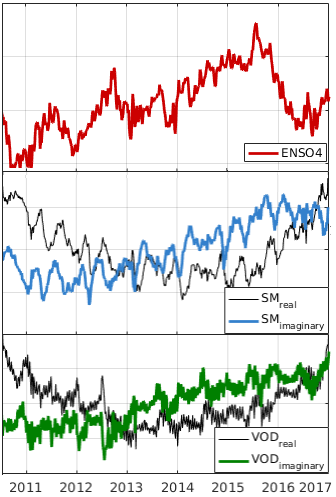}~\includegraphics[height=3.2cm]{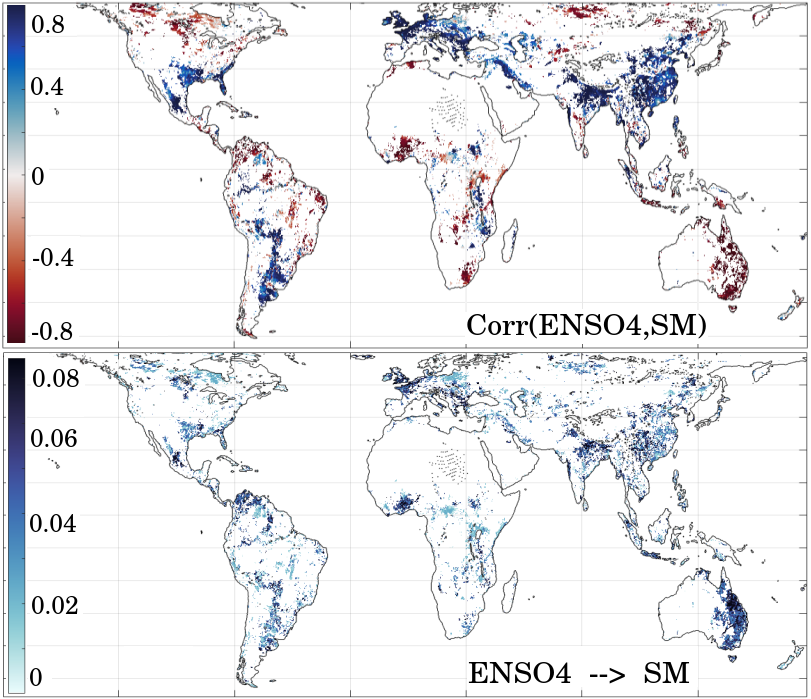}
&   \includegraphics[height=4cm]{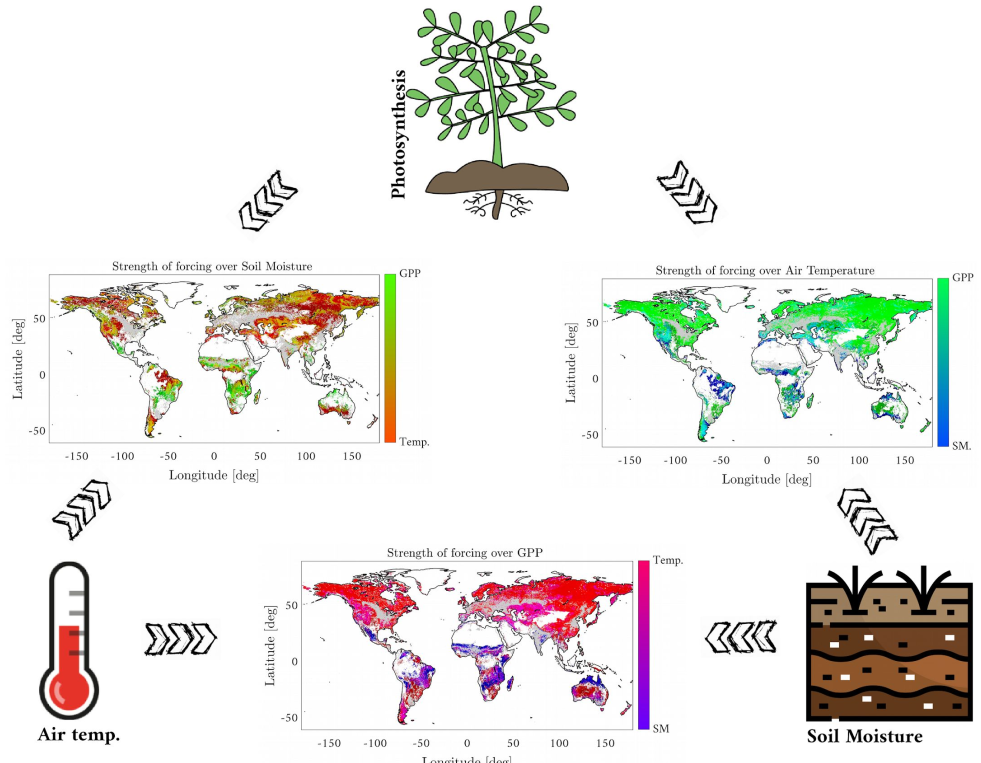} &
 \includegraphics[height=4cm]{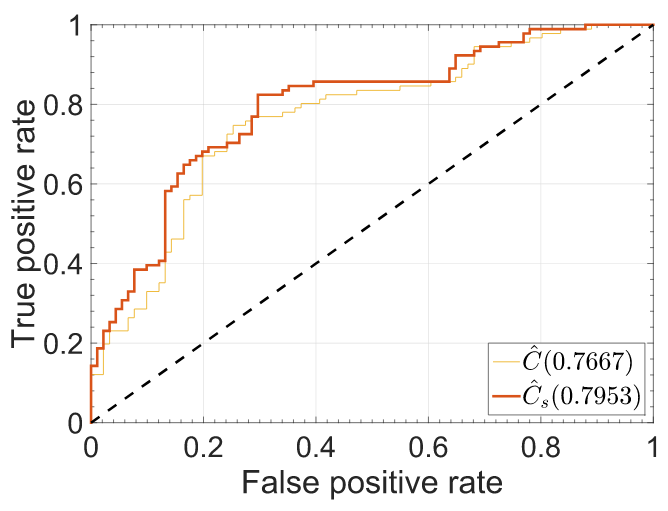}
    \end{tabular}
\end{center} 
    \caption{Examples of  causal inference approaches in remote sensing and the geosciences. 
    {\bf (a)} On the left, we show time series of ENSO4 (ENSO in region 4), which captures sea surface temperature anomalies in the central equatorial Pacific, SM and VOD, in order to explore their causal relations extracted with  nonlinear PCA in~\cite{Bueso18rock}; on the right we show the 5-day lagged correlation (top) and and causal (bottom) maps of ENSO4 and SM inter-annual components using the kernel Granger causality method in~\cite{Bueso19ci}. Results show that many of the correlations are not causal, even the highest ones ($\rho\sim0.8$), thus suggesting mere spurious associations. 
 {\bf (b)} Example of the application of the unbiased CCM in~\cite{CampsValls19agu} to derive causal relations between variables accounting for photosynthesis (gross primary productivity, GPP from FLUXCOM), temperature (T$_{\text{air}}$ from from ERA Interim) and soil moisture (SM from ESA's CCI (v 2.0)). Datacubes at $0.5^\circ$ and $8$ day spatial and temporal resolutions respectively, spanning $2001-2012$ were used. Reasonable spatial causal patterns are observed for SM and Tair on GPP; GPP drives T$_{\text{air}}$ mostly in cold ecosystems (probably due to changes in land surface albedo such as snow/ice to vegetation changes); SM is mostly controlled by T$_{\text{air}}$, which partially drives evaporation in water-limited regions; and GPP dominates SM. 
 {\bf (c)} Structural equation models in the form of additive noise model with kernels in~\cite{PerezSuay19shsic} for hypothesis testing. Assessing cause-effect relations is also possible when time is not involved. We here rely on a look-up-table (LUT) generated with by and RTM which gives the right direction of causation: state vectors (parameters) cause radiances. {The algorithms accurately detect this from pairs of data, and can be used for retrieval model-data intercomparison and RTM assessment.}
    }
    \label{fig:causalexamples}
\end{figure*}

Causal discovery methods can be divided into four main families. 
First, \emph{Granger Causality (GC)}~\cite{granger} is the most widely used approach in Earth and climate sciences to quantitatively identify relations between time series.
It tests if including past states of a variable $X$ improves the prediction of an output variable $Y$ more than considering other covariates.
GC is a linear test, but nonlinear (kernel) versions have been proposed \cite{marinazzo}. In \cite{Bueso19ci} a generalized kernel GC is presented able to discover footprints of {El Ni{\~n}o-Southern Oscillation (ENSO)} on soil moisture (SM) and vegetation optical depth (VOD) records {(see Fig.~\ref{fig:causalexamples}a)}. 
However, GC approaches have problems in nonstationary, nonlinear and deterministic relations, especially in dynamic systems with weak to moderate coupling. The second family considers nonlinear state-space methods, such as the \emph{Convergent Cross-Mapping (CCM)}~\cite{sugihara2012detecting}. CCM tries to address GC problems by reconstructing the variable's state spaces (${\bf M}_x$, ${\bf M}_y$) using time embeddings, and conclude on $X\to Y$ if points on ${\bf M}_x$ can be predicted using nearest neighbors in ${\bf M}_y$ more accurately as more points are used. However, CCM is very sensitive to noise and time series length. Recent works included bootstrap resampling to alleviate such problems, and showed good results in identifying causal links in long global records of carbon and water fluxes~\cite{CampsValls19agu}, see Fig.~\ref{fig:causalexamples}(b). The third family, collectively known as \emph{causal network learning algorithms}, heavily relies on conditional independence tests. Methods iteratively remove links between pairs of variables ($X, Y$) if they are found independently conditioned on any subset of the other variables. The PC algorithm allows to identify parents and can be flexibly implemented with different kinds of conditional independence tests, which can handle nonlinear dependencies and variables that are discrete or continuous, and univariate or multivariate. Finally, \emph{structural causal models (SCM)} are used when {\em time} is not involved or the sampling frequency is too low. SCMs  search for the causal direction within Markov equivalence classes by exploiting asymmetries between cause and effect. Additive noise models rely on the principle of independence between the cause and the generating mechanism, and have recently shown good results in remote sensing and geosciences in cases where time is not involved and only two variables are observed~\cite{PerezSuay19shsic}.

% Perspective
\subsection*{Perspectives}

While the field of machine and deep learning has traditionally progressed very rapidly, we observe that this is not the case in tackling the challenge of learning causal relations from Earth observation data.  
The role that deep learning will play on causal discovery is at best uncertain, since deep learning models mostly focus on fitting and are largely overparameterized, which is (apparently) against the causal, sparse, reasoning. Only very recently we witnessed efforts towards either incorporating or understanding deep models {\em causally}:  \cite{Bengio19}, implements a meta-learning objective that maximizes the speed of domain transfer, which under certain assumptions can be seen as a way to localize changes in causal mechanisms. In \cite{NIPS2017_7223} authors learn individual-level causal effects from observational data that can efficiently handle confounding (hidden) factors. 
Both methods are in principle well suited to the problems in remote sensing and geoscience datasets, which exhibit spatio-temporal relations to be exploited, 
but have not (so far) been considered.

Yet, we will have to face a more important challenge, the  {\em cognitive barriers}. Domain knowledge is elusive and difficult to encode, interaction between computer scientists and physicists is still a barrier, and education in synergistic concepts still needs to become a reality in the coming years. Causal inference is named to be the way to develop Earth sciences, but this will only be possible with a strong and continuous interaction between domain knowledge experts and computer scientists. 

%\begin{landscape}
%\end{landscape}

%%%%%%%%%%%%%%%%%%%%%%%%%%%%%%%%%%%%%%%%%%
\section*{Conclusion}
\label{sec:conclusions}
Six ideas, six directions where the geosciences, Earth observation and artificial intelligence still have a lot to achieve if synergystically combined. With this position paper we have provided our appreciation of research avenues that are new, refreshing and exciting for scientists willing to evolve at the interface between AI and the geosciences. We hope they will sparkle curiosity and that the community, especially the younger generations, will embrace them. 

%\iffalse  
\section*{Acknowledgment}
\par
X. Zhu is jointly supported by the European Research Council ERC under grant  ERC-2016-StG-714087, by the Helmholtz Association through the Framework of Helmholtz Artificial Intelligence Cooperation Unit (HAICU) 
and Helmholtz Excellent Professorship ``Data Science in Earth Observation - Big Data Fusion for Urban Research'' and by the German Federal Ministry of Education and Research (BMBF) in the framework of the international future AI lab ``AI4EO''. G. Camps-Valls was partly funded by the European Research Council (ERC) under the ERC-CoG-2014 SEDAL project (grant agreement 647423). 

N.\ Jacobs was partly funded by a National Science Foundation CAREER Award (IIS-1553116).

Some of the ideas in this paper originated from discussions in the first workshop of the ELLIS Program `Machine learning for Earth and Climate Science' (MFO, Germany) few days before the COVID lockdown in Europe.
%\fi

\bibliographystyle{IEEEtran}
\bibliography{bibD,gustau_refs,ribana_refs, reasoning,nbj}

\end{document}